\documentclass[journal, onecolumn]{IEEEtran}

\usepackage[utf8]{inputenc} 
\usepackage[T1]{fontenc}    
\usepackage{hyperref}       
\usepackage{url}            
\usepackage{booktabs}       
\usepackage{amsfonts}       
\usepackage{amsmath}
\usepackage{amsthm}
\usepackage{nicefrac}       
\usepackage{microtype}      
\usepackage[ruled, linesnumbered]{algorithm2e}
\usepackage{subfig}
\usepackage{graphicx}
\usepackage{float}
\usepackage{tabularx}
\usepackage{multirow}
\usepackage{makecell}
\usepackage{wrapfig}
\usepackage{booktabs}
\usepackage{array}
\usepackage[numbers]{natbib}

\newcommand{\W}{\mathcal{W}}

\newcommand{\R}{\mathbb{R}}

\DeclareMathOperator{\err}{err}
\DeclareMathOperator{\msg}{msg}

\newtheorem{theorem}{Theorem}[section]

\title{Sparse Binary Compression: Towards Distributed Deep Learning with minimal Communication}

\author{
  Felix Sattler$^1$,
  Simon Wiedemann$^2$,
  Klaus-Robert M\"uller$^3$, and
  Wojciech Samek$^4$
\thanks{$^1$F. Sattler is with the Fraunhofer Heinrich Hertz Institute, 10587 Berlin, Germany. (e-mail: felix.sattler@hhi.fraunhofer.de)}
\thanks{$^2$S. Wiedemann is with the Fraunhofer Heinrich Hertz Institute, 10587 Berlin, Germany. (e-mail: simon.wiedemann@hhi.fraunhofer.de)}
\thanks{$^3$K.-R. M\"uller is with the Technische Universit\"at Berlin, 10587 Berlin, Germany, with the Max Planck Institute for Informatics, 66123 Saarbr\"ucken, Germany, and with the Department of Brain and Cognitive Engineering, Korea University, Seoul 136-713, Korea (e-mail: klaus-robert.mueller@tu-berlin.de)}
\thanks{$^4$W. Samek is with the Fraunhofer Heinrich Hertz Institute, 10587 Berlin, Germany. (e-mail: wojciech.samek@hhi.fraunhofer.de)}}

\begin{document}

\parskip 0pt

\maketitle

\begin{abstract}
Currently, progressively larger deep neural networks are trained on ever growing data corpora. As this trend is only going to increase in the future, distributed training schemes are becoming increasingly relevant. A major issue in distributed training is the limited communication bandwidth between contributing nodes or prohibitive communication cost in general. These challenges become even more pressing, as the number of computation nodes increases. To counteract this development we propose sparse binary compression (SBC), a compression framework that allows for a drastic reduction of communication cost for distributed training. SBC combines existing techniques of communication delay and gradient sparsification with a novel binarization method and optimal weight update encoding to push compression gains to new limits. By doing so, our method also allows us to smoothly trade-off gradient sparsity and temporal sparsity to adapt to the requirements of the learning task. Our experiments show, that SBC can reduce the upstream communication on a variety of convolutional and recurrent neural network architectures by more than \emph{four} orders of magnitude without significantly harming the convergence speed in terms of forward-backward passes. For instance, we can train ResNet50 on ImageNet in the same number of iterations to the baseline accuracy, using $\times 3531$ less bits or train it to a $1\%$ lower accuracy using $\times 37208$ less bits. In the latter case, the total upstream communication required is cut from 125 \emph{terabytes} to 3.35 gigabytes for every participating client.
\end{abstract}

\section{Introduction}
\begin{figure}
\centering
\includegraphics[width=0.9\textwidth]{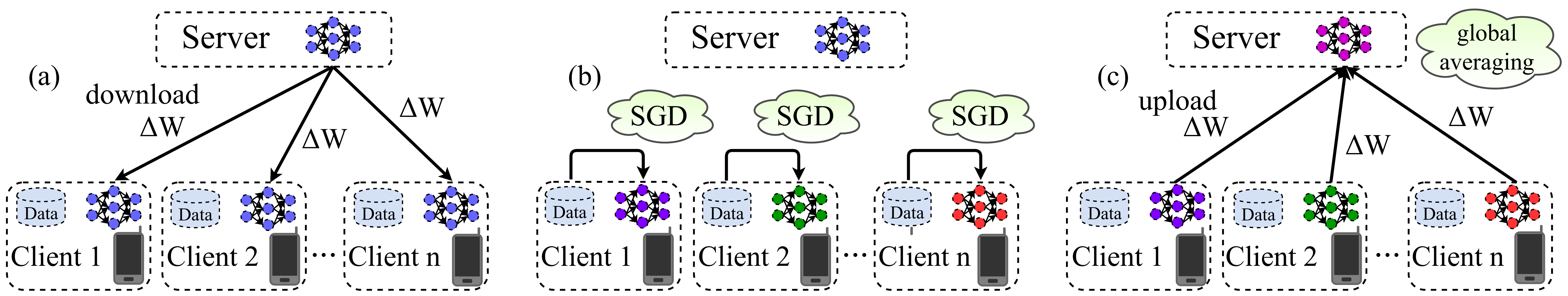}
\caption{One communication round of DSGD: a) Clients synchronize with the server. b) Clients compute a weight-update independently based on their local data. c) Clients upload their local weight-updates to the server, where they are averaged to produce the new master model.}
\label{fig:illustrate_dsgd}
\end{figure}

Distributed Stochastic Gradient Descent (DSGD) is a training setting, in which a number of clients jointly trains a deep learning model using stochastic gradient descent \cite{dean2012large}\cite{recht2011hogwild}\cite{moritz2015sparknet}. Every client holds an individual subset of the training data, used to improve the current master model. The improvement is obtained by investing computational resources to perform iterations of stochastic gradient descent (SGD). This local training produces a weight update $\Delta\W$ in every participating client, which in regular or irregular intervals (communication rounds) is exchanged 
to produce a new master model. This exchange of weight-updates can be performed indirectly via a centralized server or directly in an all-reduce operation. In both cases, all clients share the same master model after every communication round (Fig.\ \ref{fig:illustrate_dsgd}). In vanilla DSGD the clients have to communicate a full gradient update during every iteration. Every such update is of the same size as the full model, which can be in the range of gigabytes for modern architectures with millions of parameters \cite{he2016deep}\cite{huang2017densely}. Over the course of multiple hundred thousands of training iterations on big datasets the total communication for every client can easily grow to more than a \emph{petabyte}. Consequently, if communication bandwidth is limited, or communication is costly, distributed deep learning can become unproductive or even unfeasible. 

DSGD is a very popular training setting with many applications. On one end of the spectrum, DSGD can be used to greatly reduce the training time of large-scale deep learning models by introducing device-level data parallelism \cite{chilimbi2014project}\cite{zinkevich2010parallelized}\cite{xing2015petuum}\cite{li2014communication}, making use of the fact that the computation of a mini-batch gradient is perfectly parallelizable. In this setting, the clients are usually embodied by hardwired high-performance computation units (i.e.\ GPUs in a cluster) and every client performs one iteration of SGD per communication round. Since communication is high-frequent in this setting, bandwidth can be a significant bottleneck. On the other end of the spectrum DSGD can also be used to enable privacy-preserving deep learning \cite{shokri2015privacy}\cite{mcmahan2016communication}. Since the clients only ever share weight-updates, DSGD makes it possible to train a model from the combined data of all clients without any individual client having to reveal their local training data to a centralized server. In this setting the clients typically are embedded or mobile devices with low network bandwidth, intermittent network connections, and an expensive mobile data plan.   

In both scenarios, the communication cost between the individual training nodes is a limiting factor for the performance of the whole learning system. As a result of this, substantial research has gone into the effort of reducing the amount of communication necessary between the clients via lossy compression schemes \cite{li2014communication}\cite{wen2017terngrad}\cite{alistarh2017qsgd}\cite{seide20141}\cite{bernstein2018signsgd}\cite{strom2015scalable}\cite{aji2017sparse}\cite{lin2017deep}\cite{mcmahan2016communication}\cite{konevcny2016federated}.  For the synchronous distributed training scheme described above, the total amount of bits communicated by every client during training is given by
\begin{align}\label{eq:b_total}
\texttt{b}_{total} \in \mathcal{O}( \underbrace{N_{iter}\times f}_{\text{\# communication rounds}}\times ~~~\underbrace{|\Delta\W_{\neq0}| \times (\bar{\texttt{b}}_{pos}+\bar{\texttt{b}}_{val})}_{\text{\# bits per communication}}~~~\times \underbrace{K}_{\text{\# receiving nodes}})
\end{align} 
where $N_{iter}$ is the total number of training iterations (forward-backward passes) every client performs, $f$ is the communication frequency, $|\W_{\neq0}|$ is the sparsity of the weight-update, $\bar{\texttt{b}}_{pos}$, $\bar{\texttt{b}}_{val}$ are the average number of bits required to communicate the position and the value of the non-zero elements respectively and $K$ is the number of receiving nodes (if $\W$ is dense, the positions of all weights are predetermined and no position bits are required).

Existing compression schemes only focus on reducing one or two of the multiplicative components that contribute to $\mathtt{b}_{total}$. Using the systematic of equation \ref{eq:b_total}, we can group these prior approaches into three different groups:

Sparsification methods restrict weight-updates to modifying only a small subset of the parameters, thus reducing $|\Delta\W_{\neq0}|$. Strom \cite{strom2015scalable} presents an approach in which only gradients with a magnitude greater than a certain predefined threshold are sent to the server. All other gradients are aggregated into a residual. This method achieves compression rates of up to 3 orders of magnitude on an acoustic modeling task. In practice however, it is hard to choose appropriate values for the threshold, as it may vary a lot for different architectures and even different layers. Instead of using a fixed threshold to decide what gradient entries to send, Aji et al \cite{aji2017sparse} use a fixed sparsity rate. They only communicate the fraction $p$ entries of the gradient with the biggest magnitude, while also collecting all other gradients in a residual. At a sparsity rate of $p=0.001$ their method slightly degrades the convergence speed and final accuracy of the trained model. Lin et al.\ \cite{lin2017deep} present modifications to the work of Aji et al.\ which close this performance gap. These modifications include using a curriculum to slowly increase the amount of sparsity in the first couple communication rounds and applying momentum factor masking to overcome the problem of gradient staleness. They report extensive results for many modern convolutional and recurrent neural network architectures on big datasets. Using a naive encoding of the sparse weight-updates, they achieve compression rates ranging from $\times$270 to $\times$600 on different architectures, without any slowdown in convergence speed or degradation of final accuracy.

Communication delay methods try to reduce the communication frequency $f$. McMahan et al.\ \cite{mcmahan2016communication} propose Federated Averaging to reduce the communication bit-width. In Federated Averaging, instead of communicating after every iteration, every client performs multiple iterations of SGD to compute a weight-update. The authors observe that this delay of communication does not significantly harm the convergence speed in terms of local iterations and report a reduction in the number of necessary communication rounds by a factor of $\times$10 - $\times$100 on different convolutional and recurrent neural network architectures.
In a follow-up work Konevcny et al.\ \cite{konevcny2016federated} combine this communication delay with random sparsification and probabilistic quantization. They restrict the clients to learn random sparse weight-updates or force random sparsity on them afterwards ("structured" vs "sketched" updates) and combine this sparsification with probabilistic quantization. While their method also combines communication delay with (random) sparsification and quantization, and achieves good compression gains for one particular CNN and LSTM model, it also causes a major drop in convergence speed and final accuracy.  

Dense quantization methods try to reduce the amount of value bits $\bar{\texttt{b}}_{val}$. Wen et al.\ propose TernGrad \cite{wen2017terngrad}, a method to stochastically quantize gradients to ternary values. This achieves a moderate compression rate of $\times$16, while accuracy drops noticeably on big modern architectures. The authors prove the convergence of their method under the assumption of bounded gradients.
Alistarh et al.\ \cite{alistarh2017qsgd}, explore the trade-off between model accuracy and gradient precision. They prove information theoretic bounds on the compression rate achievable by dense quantization and propose QSGD, a family of compression schemes with convergence guarantees. Other authors experiment with 1-bit quantization schemes: Seide et al.\ \cite{seide20141} show empirically that it is possible to quantize the weight-updates to 1 bit without harming convergence speed, if the quantization errors are accumulated. Bernstein et al.\ \cite{bernstein2018signsgd} propose signSGD, a distributed training scheme in which every client quantizes the gradients to binary signs and the server aggregates the gradients by means of a majority vote.
In general of course, dense quantization can only achieve a maximum compression rate of $\times$32. 


\section{Sparse Binary Compression}
\begin{figure}[H]
\centering

\includegraphics[width=0.9\textwidth]{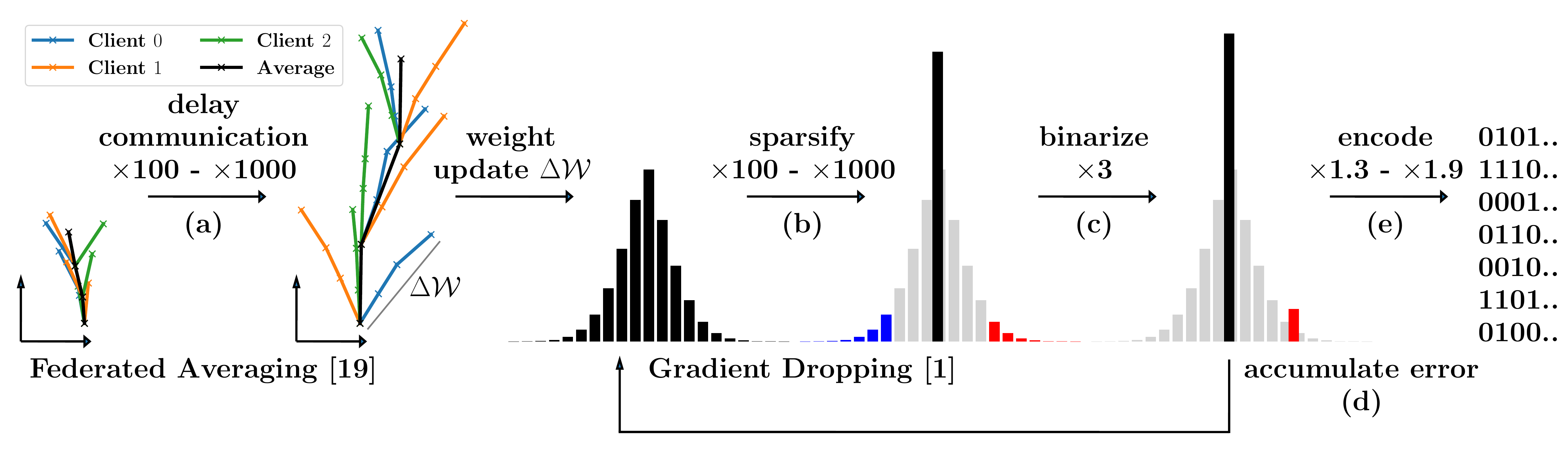}
\caption{Step-by-step explanation of techniques used in Sparse Binary Compression: (a) Illustrated is the traversal of the parameter space with regular DSGD (left) and Federated Averaging (right). With this form of communication delay, a bigger region of the loss surface can be traversed, in the same number of communication rounds. That way compression gains of up to $\times 1000$ are possible. After a number of iterations, the clients communicate their locally computed weight-updates. (b) Before communication, the weight-update is first sparsified, by dropping all but the fraction $p$ weight-updates with the highest magnitude. This achieves up to $\times 1000$ compression gain. (c) Then the sparse weight-update is binarized for an additional compression gain of approximately  $\times 3$. (d) Finally, we optimally encode the positions of the non-zero elements, using Golomb encoding. This reduces the bit size of the compressed weight-update by up to another $\times 2$ compared to naive encoding.}
\label{fig:explain}
\end{figure}

We propose Sparse Binary Compression (cf.\ Figure \ref{fig:explain}), to drastically reduce the number of communicated bits in distributed training. SBC makes use of multiple techniques \emph{simultaneously\footnote{To clarify, we have put our contributions in emphasis.}} to reduce all multiplicative components of equation \eqref{eq:b_total}. 

In the following $\W$ will refer to the entirety of neural network parameters, while $W\in\W$ will refer to one specific tensor of weights. Arithmetic operations on $\W$ are to be understood componentwise.

\begin{figure*}[bbb!]
\begin{minipage}[t]{3.5in}

\begin{algorithm}[H] \label{alg:DSGD}
\caption{Synchronous Distributed Stochastic Gradient Descent (DSGD)}
\DontPrintSemicolon
\textbf{input:} initial parameters $\W$\\
\textbf{outout:} improved parameters $\W$\\
\textbf{init:} all clients $C_i$ are initialized with the same parameters $\W_i\leftarrow \W$, the initial global weight-update and the residuals are set to zero $\Delta \W,\mathcal{R}_i\leftarrow 0$\\
\For{$t=1,..,T$}{
\For{$i \in I_t\subseteq \{1,..,M\}$ \textbf{in parallel}}{
\underline{Client $C_i$ does:}\\
\textbullet~ $\text{msg} \leftarrow \text{download}_{S \rightarrow C_i}(\text{msg})$\\
\textbullet~ ${\Delta \W} \leftarrow \text{decode}(\text{msg})$\\
\vspace{8pt}
\textbullet~ $\W_i \leftarrow \W_i + {\Delta \W}$ \\
\textbullet~ $\Delta \W_i \leftarrow \mathcal{R}_i+\text{SGD}_n(\W_i, D_i)-\W_i$ \\
\textbullet~ $\Delta \W_i^* \leftarrow \text{compress}(\Delta\W_i)$\\
\textbullet~ $\mathcal{R}_i \leftarrow \Delta \W_i - \Delta \W_i^*$ \\
\vspace{8pt}
\textbullet~ $\text{msg}_i \leftarrow \text{encode}(\Delta \W_i^*)$\\
\textbullet~ $\text{upload}_{C_i \rightarrow S}(\text{msg}_i)$
}
\underline{Server $S$ does:}\\
\textbullet~ $\text{gather}_{C_i\rightarrow S}(\Delta \W_i^*),~i\in I_t$\\
\textbullet~ $\Delta \W \leftarrow \frac{1}{|I_t|}\sum_{i\in I_t}\Delta \W^*_i$\\
\textbullet~ $\W\leftarrow \W+\Delta \W$\\
\textbullet~ $\text{broadcast}_{S\rightarrow C_i}(\Delta \W),~i=1,..,M$
}
\Return $\W$

\end{algorithm}
\end{minipage}
\hfill
\begin{minipage}[t]{3.5in}

\begin{algorithm}[H] \label{alg:SBC}
\caption{Sparse Binary Compression}
\DontPrintSemicolon
\textbf{input:} tensor $\Delta W$, sparsity $p$\\
\textbf{output:} sparse tensor $\Delta W^*$\\
\textbullet~ $\text{val}^+ \leftarrow \text{top}_{p\%}(\Delta W)$; $\text{val}^- \leftarrow \text{top}_{p\%}(-\Delta W)$\\
\textbullet~ $\mu^+ \leftarrow \text{mean}(\text{val}^+)$; $\mu^- \leftarrow \text{mean}(\text{val}^-)$\\
\uIf{$\mu^+ \geq \mu^-$}{
\Return $\Delta W^* \leftarrow \mu^+(W \geq \min(\text{val}^+))$ \\
}
\Else{
\Return $\Delta W^* \leftarrow -\mu^-(W \leq -\min(\text{val}^-))$ \\
}
\end{algorithm}

\begin{algorithm}[H] \label{alg:encode}
\caption{Golomb Position Encoding}
\DontPrintSemicolon
\textbf{input:} sparse tensor $\Delta W^*$, sparsity $p$\\
\textbf{output:} binary message msg\\
\textbullet~ $\mathcal{I} \leftarrow \Delta W^*[:]_{\neq 0}$\\
\textbullet~ $\mathbf{b}^* \leftarrow 1+\lfloor \log_2(\frac{\log(\phi-1)}{\log(1-p)})\rfloor$\\
\For{$i=1,..,|\mathcal{I}|$}{
\textbullet~ $d \leftarrow \mathcal{I}_{i}-\mathcal{I}_{i-1}$\\
\textbullet~ $q \leftarrow  {(d-1)} \text{ div } {2^{\mathtt{b}^*}}$\\
\textbullet~ $r \leftarrow {(d-1)} \text{ mod } {2^{\mathtt{b}^*}}$\\
\textbullet~ msg.add($\underbrace{1, .., 1}_{q \text{ times}}$, 0, binary$_{\mathtt{b}^*}(r)$)\\
}
\Return msg
\end{algorithm}
\end{minipage}
\end{figure*}

\textbf{Communication Delay, Fig.\ \ref{fig:explain} (a)}:
We use communication delay, proposed by \cite{mcmahan2016communication}, to introduce temporal sparsity into DSGD. Instead of communicating gradients after every local iteration, we allow the clients to compute more informative updates by performing multiple iterations of SGD. These generalized weight-updates are given by
\begin{align*}
\Delta \mathcal{W}_i =\text{SGD}_n(\W_i,D_i)-\W_i
\end{align*}
where $\text{SGD}_n(\W_i,D_i)$ refers to the set of weights obtained by performing $n$ iterations of stochastic gradient descent on $\W_i$, while sampling mini-batches from the i-th client's training data $D_i$. Empirical analysis by \cite{mcmahan2016communication} suggests that communication can be delayed drastically, with only marginal degradation of accuracy. For $n=1$ we obtain regular DSGD. 

\textbf{Sparse Binarization, Fig.\ \ref{fig:explain} (b), (c)}:
Following the works of \cite{lin2017deep}\cite{strom2015scalable}\cite{shokri2015privacy}\cite{aji2017sparse} we use the magnitude of an individual weight within a weight-update as a heuristic for it's importance. First, we set all but the fraction $p$ biggest and fraction $p$ smallest weight-updates to zero. Next, {we compute the mean of all remaining positive and all remaining negative weight-updates independently. \emph{If the positive mean $\mu^+$ is bigger than the absolute negative mean $\mu^-$, we set all negative values to zero and all positive values to the positive mean and vice versa}. The method is illustrated in figure \ref{fig:explain} and formalized in algorithm \ref{alg:SBC}.
Finding the fraction $p$ smallest and biggest values in a vector $W$ requires $\mathcal{O}(|W|\log(2p|W|))$ operations, where $|W|$ refers to the number of elements in $W$. As suggested in \cite{lin2017deep}, we can reduce the computational cost of this sorting  operation, by randomly subsampling from $W$. However this comes at the cost of introducing (unbiased) noise in the amount of sparsity. Luckily, in our approach communication rounds (and thus compressions) are relatively infrequent, which helps to marginalize the overhead of the sparsification. \emph{Quantizing the non-zero elements of the sparsified weight-update to the mean reduces the required value bits $\bar{\texttt{b}}_{val}$ from 32 to 0}. This translates to a reduction in communication cost by a factor of around $\times3$. We can get away with averaging out the non-zero weight-updates because they are relatively homogeneous in value and because we accumulate our compression errors as described in the next paragraph. 

Although other methods, like TernGrad \cite{wen2017terngrad} also combine sparsification and quantization of the weight-updates, none of these methods work with sparsity rates as high as ours (see Table \ref{tab:theoretical}). 

\textbf{Residual Accumulation, Fig.\ \ref{fig:explain} (d)}:
It is well established (see \cite{lin2017deep}\cite{strom2015scalable}\cite{aji2017sparse}\cite{seide20141}) that the convergence in sparsified DSGD can be greatly accelerated by accumulating the error that arises from only sending sparse approximations of the weight-updates. After every communication round, the residual is updated via
\begin{align}
\mathcal{R}_\tau = \sum_{t=1}^\tau (\Delta \W_t - \Delta \W^*_t) =  \mathcal{R}_{\tau-1}+\Delta \W_\tau - \Delta \W^*_\tau .
\label{eq:residual}
\end{align}
Error accumulation has the great benefit that no gradient information is lost (it may only become outdated or "stale"). In the context of pure sparsification residual accumulation can be interpreted to be equivalent to increasing the batch size for individual parameters \cite{lin2017deep}. Moreover, we can show:
\begin{theorem}
\label{theo:1}
Let $\Delta W_1,..,\Delta W_T\in\R^n$ be (flattened) weight-updates, computed by one client in the first $T$ communication rounds. Let $\Delta W_1^*,..,\Delta W_{T-1}^* \in \mathcal{S}$ be the actual weight-updates, transferred in the previous rounds (restricted to some subspace $\mathcal{S}$) and $\mathcal{R}_\tau$ be the content of the residual at time $\tau$ as in \eqref{eq:residual}. Then the orthogonal projection
\begin{align}
v = \text{Proj}_\mathcal{S}(\mathcal{R}_{T-1}+\Delta W_T)
\end{align}
uniquely minimizes the accumulated error
\begin{align}
\err(\Delta W_T^*)=\|\sum_{t=1}^T (\Delta W_t-\Delta W^*_t)\|
\end{align} 
in $\mathcal{S}$. \emph{(Proof in Supplement.)}
\end{theorem}

That means that the residual accumulation keeps the compressed optimization path as close as possible to optimization path taken with non-compressed weight-updates.

\textbf{Optimal Position Encoding, Fig.\ \ref{fig:explain} (e)}:
To communicate a set of sparse binary tensors produced by SGC, we only need to transfer the positions of the non-zero elements in the flattened tensors, along with one mean value ($\mu^+$ or $\mu^-$) per tensor. Instead of communicating the absolute non-zero positions it is favorable to only communicate the distances between all non-zero elements. Under the simplifying assumption that the sparsity pattern is random for every weight-update, it is easy to show that these distances are geometrically distributed with success probability $p$ equal to the sparsity rate. Therefore, as previously done by \cite{strom2015scalable}, we can optimally encode the distances using the Golomb code \cite{golomb1966run}. Golomb encoding reduces the average number of position bits to 
\begin{align}
\bar{\texttt{b}}_{pos} = \mathbf{b}^*+\frac{1}{1-(1-p)^{2^{\mathbf{b}^*}}},
\end{align}
with $\mathbf{b}^*=1+\lfloor \log_2(\frac{\log(\phi-1)}{\log(1-p)})\rfloor$ and $\phi=\frac{\sqrt{5}+1}{2}$ being the golden ratio. For a sparsity rate of i.e.\ $p=0.01$, we get $\bar{\texttt{b}}_{pos}=8.38$, which translates to $\times 1.9$ compression, compared to a naive distance encoding with 16 fixed bits. While the overhead for encoding and decoding makes it unproductive to use Golomb encoding in the situation of \cite{strom2015scalable}, this overhead becomes negligible in our situation due to the infrequency of weight-update exchange resulting from communication delay. The encoding scheme is given in algorithm \ref{alg:encode}, while the decoding scheme can be found in the supplement.

\textbf{Momentum Correction, Warm-up Training and Momentum Masking:}
Lin et al.\ \cite{lin2017deep} introduce multiple minor modifications to the vanilla Gradient Dropping method, to improve the convergence speed. We adopt momentum masking, while momentum correction is implicit to our approach. For more details on this we refer to the supplement.

\ \\
Our proposed method is described in Algorithms \ref{alg:DSGD}, \ref{alg:SBC} and \ref{alg:encode}. Algorithm \ref{alg:DSGD} describes how compression and residual accumulation can be introduced into DSGD. Algorithm \ref{alg:SBC} describes our compression method. Algorithm \ref{alg:encode} describes the Golomb encoding. Table \ref{tab:theoretical} compares  theoretical asymptotic compression rates of different popular compression methods.

\begin{table}[H]
\centering
{\renewcommand{\arraystretch}{1.1}
\scalebox{1.1}{
\begin{tabular}{|lc|c|c|c|c|c|}

\hline
 \multicolumn{2}{|c|}{\makecell{${\text{Total Bits}}$ $=$}}  & Baseline & \makecell{SignSGD \cite{bernstein2018signsgd},TernGrad \cite{wen2017terngrad},\\QSGD \cite{alistarh2017qsgd},  1-bitSGD \citep{seide20141}} & \makecell{Gradient Dropping \citep{aji2017sparse},\\  DGC \cite{lin2017deep}  } & \makecell{Federated\\Averaging \cite{mcmahan2016communication}}  & \makecell{Sparse Binary\\Compression}  \\
\hline
& Temporal Sparsity & 100\% & 100\% & 100\% & \textbf{0.1\% - 10\%} & \textbf{0.1\% - 10\%} \\
$\times$ & Gradient Sparsity & 100\% & 100\% & \textbf{0.1\%}  &  100\% & \textbf{0.1\% - 10\%} \\
\multirow{2}{*}{$\times \sum$} & Value Bits  & $32$ & \textbf{1 - 8} & $32$  & $32$ & $\mathbf{0}$ \\
& Position Bits & $0$ & $0$ & $16$ & $0$ & $\mathbf{8}$ - $\mathbf{14}$\\
\hline
 \multicolumn{2}{|c|}{Compression Rate}& $\mathbf{\times 1}$ & $\mathbf{\times 4}$ - $\mathbf{\times 32}$ & $\mathbf{\times 666}$  & \makecell{$\mathbf{\times 10}$ - $\mathbf{\times 1000}$} & - $\mathbf{\times 40000}$\\
\hline
\end{tabular}
}
}
\vspace{5pt}
\caption{Theoretical asymptotic compression rates for different compression methods broken down into components. Only SBC reduces all multiplicative components of the total bitsize (cf.\ eq.\ \ref{eq:b_total}).}
\label{tab:theoretical}
\end{table}

\section{Temporal vs Gradient Sparsity}
\label{sec:ts_vs_gs}

Communication constraints can vary heavily between learning tasks and may not even be consistent throughout one distributed training session. Take, for example, a set of mobile devices, jointly training a model with privacy-preserving DSGD \cite{mcmahan2016communication}\cite{shokri2015privacy}. Part of the day, the devices might be connected to wifi, enabling them to frequently exchange weight-updates (still with as small of a bit size as possible), while at other times an expensive or limited mobile plan might force the devices to delay their communication. Practical methods should be able to adapt to these fluctuations in communication constraints.
We take an holistic view towards communication efficient distributed deep learning by observing that communication delay and weight-update compression can be viewed as two types of sparsity, that both affect the total number of parameters, updated throughout training, in a multiplicative way (cf. fig. \ref{fig:sparsity_plane}). While compression techniques sparsify individual gradients, communication delay sparsifies the gradient information in time.

\begin{wrapfigure}{R}{0.35\textwidth}
\includegraphics[width=0.35\textwidth]{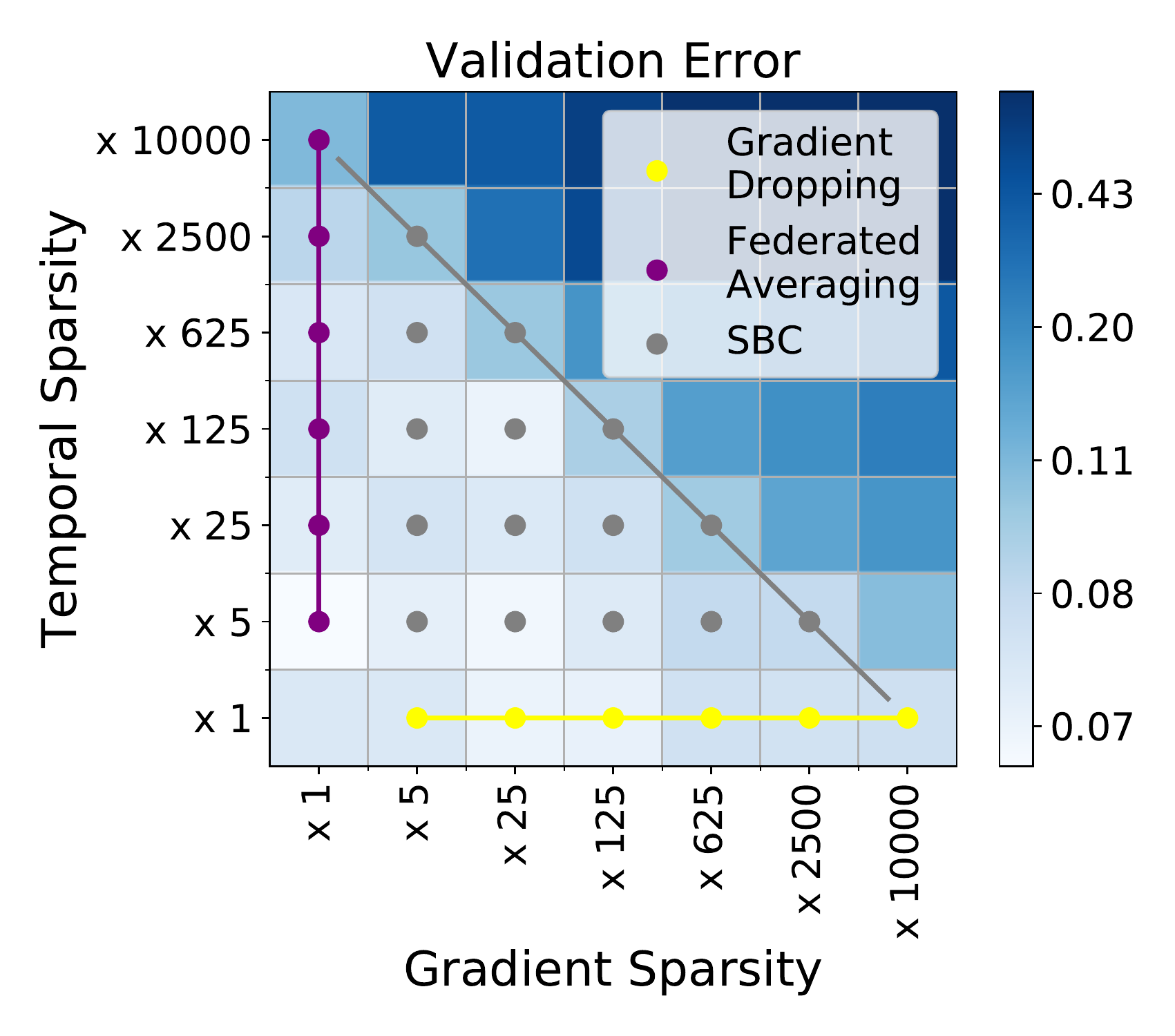}
\caption{Validation Error for ResNet32 trained on CIFAR at different levels of temporal and gradient sparsity (the error is color-coded, brighter means lower error).}
\label{fig:sparsity_plane}
\end{wrapfigure}

Our method is unique in the sense that it allows us to smoothly trade of these two types of sparsity against one another. Figure \ref{fig:sparsity_plane} shows validation errors for ResNet32 trained on CIFAR (model specification in section \ref{sec:models}) for 60000 iterations at different levels of temporal and gradient sparsity. Along the off-diagonals of the matrix, the total sparsity, defined as the product of temporal and gradient sparsity, remains constant. We observe multiple things:
1.) The validation error remains more or less constant along the off-diagonals of the matrix.
2.) Federated Averaging (purple) and Gradient Dropping/ DGC (yellow) are just lines in the two-dimensional space of possible compression methods.
3.) There exists a roughly triangular area of approximately constant error, \emph{optimal compression methods lie along the hypotenuse of this triangle}. We find this behavior consistently across different model architectures, more examples can be found in the supplement.
These results indicate, that there exists a fixed communication budged in DSGD, necessary to achieve a certain accuracy. Figure \ref{fig:total_sparsity} shows validation errors for the same ResNet32 model trained on CIFAR at different levels of total sparsity and different numbers of training iterations. We observe two distinct phases during training: In the beginning (iterations 0 - 30000), when training is performed using a high learning-rate, sparsified methods consistently achieve the lowest error and temporally sparsified DSGD tends to outperform purely gradient sparsified DSGD at all sparsity levels. After the learning rate is decreased by a factor of 10 in iteration 30000, this behavior is reversed and gradient sparsification methods start to perform better that temporally sparsified methods. These results highlight, that an optimal compression strategy, needs to be able to adapt temporal and gradient sparsity, not only based on the learning task, but also on the current learning stage. Such an adaptive sparsity can be integrated seamlessly into our SBC framework.

\begin{figure}[H]
\centering
\includegraphics[width=0.75\textwidth]{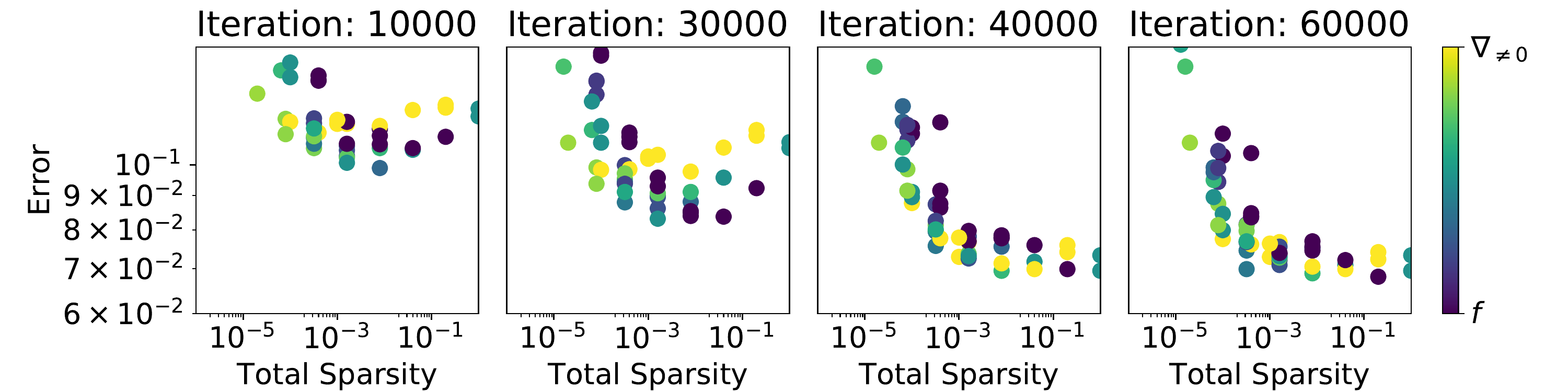}
\caption{Classification error at different levels of total sparsity and different numbers of training iterations. Purple dots represent purely temporal sparsified SGD, while yellow dots represent purely gradient sparsified SGD. For hybrid methods, the color-code interpolates between purple and yellow.}
\label{fig:total_sparsity}
\end{figure}

\section{Experiments}
\label{sec:experiments}

\subsection{Networks and Datasets}
\label{sec:models}
We evaluate our method on commonly used convolutional and recurrent neural networks with millions of parameters, which we train on well-studied data sets that contain up to multiple millions of samples. Throughout all of our experiments,  we fix the number of clients to 4 and split the training data among the clients in a balanced way (the number of training samples and their distribution is homogenous among the clients). 

\textbf{Image Classification:} We run experiments for LeNet5-Caffe\footnote{A  modified  version  of  LeNet5  from  \cite{lecun1998gradient} (see supplement).} on MNIST \cite{lecun1998mnist}, ResNet32 \cite{he2016deep} on CIFAR-10 \cite{krizhevsky2014cifar} and ResNet50 on ILSVRC2012 (ImageNet) \cite{deng2009imagenet}. We split the training data randomly into 4 shards of equal size, and assign one shard to every one of the 4 clients. The MNIST model is trained using the Adam optimizer \cite{kingma2014adam}, while the other models are trained using momentum SGD. Learning rate, weight intitiallization and data augmentation are as in the respective papers. 

\textbf{Language Modeling:} We experiment with multilayer sequence-to-sequence LSTM models as described in \cite{zaremba2014recurrent} on the Penn Treebank corpus (PTB) \cite{marcus1993building} and Shakespeare dataset for next-word and next-character prediction. The PTB dataset consists of a sequence 923000 training, and 82000 validation words. We restrict the vocabulary to the most common 10000 words, plus an additional token for all words that are less frequent and train a two-layer LSTM model with 650 hidden units ("WordLSTM"). The Shakespeare dataset consists of the complete works of William Shakespeare \cite{shakespeare2014complete} concatenated to a sequence of 5072443 training and 105675 test characters. The number of different characters in the dataset is 98. We train the two-layer "CharLSTM" with 200 hidden units. For both datasets, we split the sequences of training symbols each into four subsequences of equal length and assign every client one of these subsequences. 

While the models we use in our experiments do not fully achieve state-of-the-art results on the respective tasks and datasets, they are still sufficient for the purpose of evaluating our compression method and demonstrate, that our method works well with common regularization techniques such as batch normalization \cite{ioffe2015batch} and dropout \cite{srivastava2014dropout}. A complete description of models and hyperparameters can be found in the supplement.

\subsection{Results}
We experiment with three configurations of our method: SBC (1) uses no communication delay and a gradient sparsity of 0.1\%, SBC (2) uses 10 iterations of communication delay and 1\% gradient sparsity and  SBC (3) uses 100 iterations of communication delay and 1\% gradient sparsity. Our decision for these points on the 2D grid of possible configurations is somewhat arbitrary. The experiments with SBC (1) serve the purpose of enabling us to directly compare our 0-value-bit quantization to the 32-value-bit Gradient Dropping (with momentum correction and momentum factor masking \citep{lin2017deep}).

\begin{table}[H]
\centering
{\renewcommand{\arraystretch}{1.1}
\scalebox{1.05}{
\begin{tabular}{|c|c|c|c|c|c|c|c|c|}
\hline
\multicolumn{2}{|c|}{Method} & Baseline  & \makecell{Gradient\\Droping \citep{aji2017sparse}} & \makecell{Federated\\Averaging \cite{mcmahan2016communication}} &  SBC (1) & SBC (2) & SBC (3)\\
\hline
\hline
\multirow{2}{*}{\makecell{LeNet5-Caffe\\@MNIST}}& Accuracy & 0.9946  & 0.994 & 0.994 & 0.994 & 0.994 & 0.991\\
& Compression & $\times1$  & $\times 634$ & $\times 500$ & $\times 2071$ & $\times 3491$ & $\mathbf{\times 24935}$\\
\hline
\hline
\multirow{2}{*}{\makecell{ResNet32\\@CIFAR}}& Accuracy & 0.926 & 0.927 & 0.919 & 0.923 & 0.919 & 0.922 \\
& Compression & $\times 1$  & $\times 604$ & $\times 1000$ & $\times 1530$ & $\times 3430$ & $\mathbf{\times 32300}$\\
\hline
\hline
\multirow{2}{*}{\makecell{ResNet50\\@ImageNet}}& Accuracy & 0.737  & 0.739  & 0.724 & 0.735 & 0.737 & 0.728 \\
& Compression & $\times 1$  & $\times 601$ & $\times 1000$ & $\times 2569$ & $\times{3531}$ & $\mathbf{\times 37208}$ \\
\hline
\hline
\multirow{2}{*}{\makecell{WordLSTM\\@PTB}}& Perplexity & 89.16  & 89.39 & 88.59 & 89.32 & 88.47 & 89.31 \\
& Compression & $\times 1$  & $\times 665$ & $\times 1000$ & $\times 2521$ & $\times 3460$ & $\mathbf{\times 34905}$ \\
\hline
\hline
\multirow{2}{*}{\makecell{CharLSTM\\@Shakespeare}}& Perplexity & 3.635  & 3.639 & 3.904 & 3.671 & 3.782 & 4.072 \\
& Compression & $\times 1$  & $\times 660$ & $\times 1000$ & $\times 2572$ & $\times 3958$ & $\mathbf{\times 35201}$ \\
\hline
\hline
\end{tabular}
}
}
\vspace{5pt}
\caption{Final accuracy/perplexity and compression rate for different compression schemes.}
\label{tab:results}
\end{table}

Table \ref{tab:results} lists compression rates and final validation accuracies achieved by different compression methods, when applied to the training of neural networks on 5 different datasets. The number of iterations (forward-backward-passes) is held constant for all methods. On all benchmarks, our methods perform comparable to the baseline, while communicating significantly less bits.

Figure \ref{fig:results_imagenet} shows convergence speed in terms of iterations (left) and communicated bits (right) respectively for ResNet50 trained on ImageNet. The convergence speed is only marginally affected, by the different compression methods. In the first 30 epochs SBC (3) even achieves the highest accuracy, using about $\times 37000$ less bits than the baseline. In total, SBC (3) reduces the upstream communication on this benchmark from 125 \emph{terabytes} to 3.35 gigabytes for every participating client. After the learning rate is lowered in epochs 30 and 60 progress slows down for SBC (3) relative to the methods which do not use communication delay. In direct comparison SBC (1) performs very similar to Gradient Dropping, while using about $\times 4$ less bits (that is $\times 2569$ less bits than the baseline). 
\begin{figure}[H]
\centering
\includegraphics[width=\textwidth]{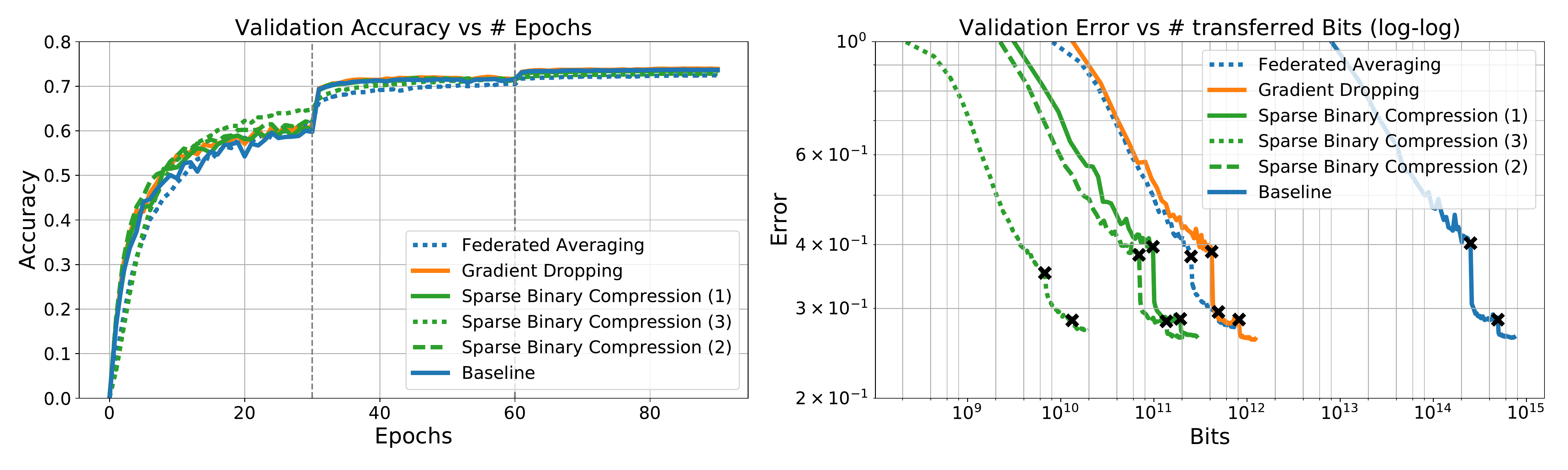}
\caption{Left: Top-1 validation accuracy vs number of epochs. Right: Top-1 validation error vs number of transferred bits (log-log). Epochs 30 and 60 at which the learning rate is reduced are marked in the plot. ResNet50 trained on ImageNet.}
\label{fig:results_imagenet}
\end{figure}
Figure \ref{fig:results_ptb} shows convergence speed in terms of iterations (left) and communicated bits (right) respectively for WordLSTM trained on PTB. While Federated Averaging and SBC (3) initially slow down convergence in terms of iterations, all models converge to approximately the same perplexity after around 60 epochs. Throughout all experiments, SBC (2) performs very similar to SBC (1) in terms of convergence speed and final accuracy, while maintaining a compression-edge of about $\times 1.3$ - $\times 2.2$.

\begin{figure}[H]
\centering
\includegraphics[width=\textwidth]{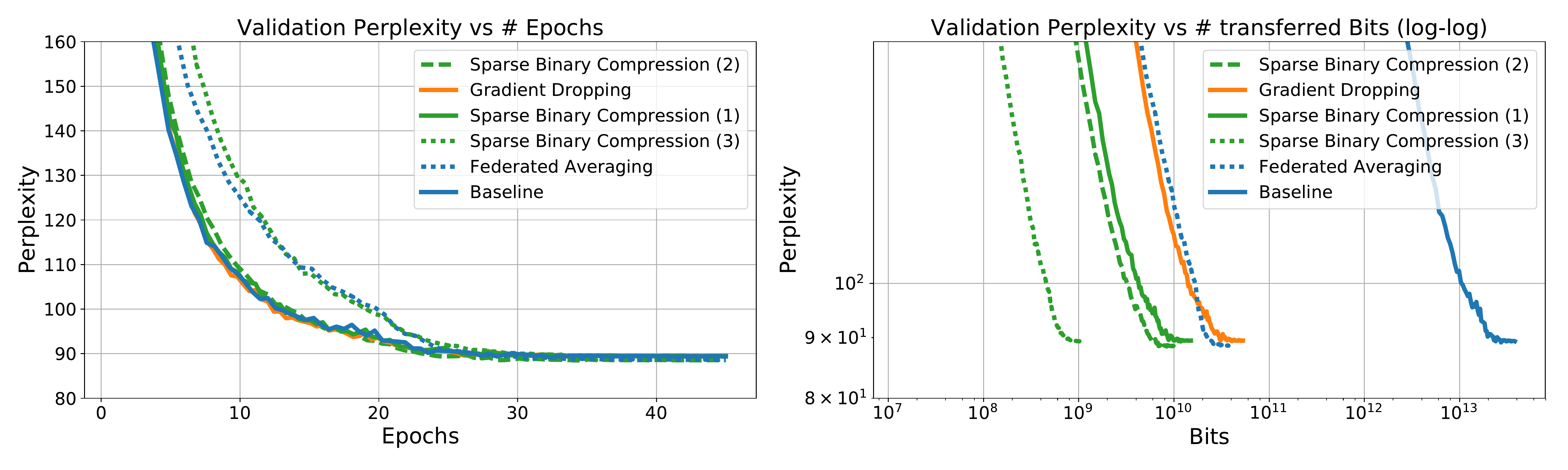}
\caption{Perplexity vs number of epochs and number of transferred bits. WordLSTM on PTB.}
\label{fig:results_ptb}
\end{figure}

\section{Conclusion}
The gradient information for training deep neural networks with SGD is highly redundant (see e.g.\ \cite{lin2017deep}). We exploit this fact to the extreme by \emph{combining} 3 powerful compression strategies and are able to achieve compression gains of up to \emph{four orders} of magnitude with only a slight decrease in accuracy. We show through experiments that communication delay and gradient sparsity can be viewed as two independent types of sparsity, that have similar effects on the convergence speed, when introduced into distributed SGD. We would like to highlight, that in no case we did modify the hyperparameters of the respective baseline models to accommodate our method. This demonstrates that our method is easily applicable. Note however that an extensive hyperparameter search could further improve the results. Furthermore, our findings in section \ref{sec:ts_vs_gs} indicate that even higher compression rates are possible if we adapt both the kind of sparsity and the sparsity rate to the current training phase. It remains an interesting direction of further research to identify heuristics and theoretical insights that can exploit these fluctuations in the training statistics to guide sparsity towards optimality.  

\section{Acknowledgements}
This work was supported by the Fraunhofer Society through the MPI-FhG collaboration project "Theory \& Practice for Reduced Learning Machines". This work was also supported by the German Ministry for Education and Research as Berlin Big Data Center under Grant 01IS14013A.

\bibliographystyle{IEEEtran}
\bibliography{sample.bib}

\begin{thebibliography}{10}
\providecommand{\url}[1]{#1}
\csname url@samestyle\endcsname
\providecommand{\newblock}{\relax}
\providecommand{\bibinfo}[2]{#2}
\providecommand{\BIBentrySTDinterwordspacing}{\spaceskip=0pt\relax}
\providecommand{\BIBentryALTinterwordstretchfactor}{4}
\providecommand{\BIBentryALTinterwordspacing}{\spaceskip=\fontdimen2\font plus
\BIBentryALTinterwordstretchfactor\fontdimen3\font minus
  \fontdimen4\font\relax}
\providecommand{\BIBforeignlanguage}[2]{{%
\expandafter\ifx\csname l@#1\endcsname\relax
\typeout{** WARNING: IEEEtran.bst: No hyphenation pattern has been}%
\typeout{** loaded for the language `#1'. Using the pattern for}%
\typeout{** the default language instead.}%
\else
\language=\csname l@#1\endcsname
\fi
#2}}
\providecommand{\BIBdecl}{\relax}
\BIBdecl

\bibitem{dean2012large}
J.~Dean, G.~Corrado, R.~Monga, K.~Chen, M.~Devin, M.~Mao, A.~Senior, P.~Tucker,
  K.~Yang, Q.~V. Le \emph{et~al.}, ``Large scale distributed deep networks,''
  in \emph{Advances in neural information processing systems}, 2012, pp.
  1223--1231.

\bibitem{recht2011hogwild}
B.~Recht, C.~Re, S.~Wright, and F.~Niu, ``Hogwild: A lock-free approach to
  parallelizing stochastic gradient descent,'' in \emph{Advances in neural
  information processing systems}, 2011, pp. 693--701.

\bibitem{moritz2015sparknet}
P.~Moritz, R.~Nishihara, I.~Stoica, and M.~I. Jordan, ``Sparknet: Training deep
  networks in spark,'' \emph{arXiv preprint arXiv:1511.06051}, 2015.

\bibitem{he2016deep}
K.~He, X.~Zhang, S.~Ren, and J.~Sun, ``Deep residual learning for image
  recognition,'' in \emph{Proceedings of the IEEE conference on computer vision
  and pattern recognition}, 2016, pp. 770--778.

\bibitem{huang2017densely}
G.~Huang, Z.~Liu, K.~Q. Weinberger, and L.~van~der Maaten, ``Densely connected
  convolutional networks,'' in \emph{Proceedings of the IEEE conference on
  computer vision and pattern recognition}, vol.~1, no.~2, 2017, p.~3.

\bibitem{chilimbi2014project}
T.~M. Chilimbi, Y.~Suzue, J.~Apacible, and K.~Kalyanaraman, ``Project adam:
  Building an efficient and scalable deep learning training system.'' in
  \emph{OSDI}, vol.~14, 2014, pp. 571--582.

\bibitem{zinkevich2010parallelized}
M.~Zinkevich, M.~Weimer, L.~Li, and A.~J. Smola, ``Parallelized stochastic
  gradient descent,'' in \emph{Advances in neural information processing
  systems}, 2010, pp. 2595--2603.

\bibitem{xing2015petuum}
E.~P. Xing, Q.~Ho, W.~Dai, J.~K. Kim, J.~Wei, S.~Lee, X.~Zheng, P.~Xie,
  A.~Kumar, and Y.~Yu, ``Petuum: A new platform for distributed machine
  learning on big data,'' \emph{IEEE Transactions on Big Data}, vol.~1, no.~2,
  pp. 49--67, 2015.

\bibitem{li2014communication}
M.~Li, D.~G. Andersen, A.~J. Smola, and K.~Yu, ``Communication efficient
  distributed machine learning with the parameter server,'' in \emph{Advances
  in Neural Information Processing Systems}, 2014, pp. 19--27.

\bibitem{shokri2015privacy}
R.~Shokri and V.~Shmatikov, ``Privacy-preserving deep learning,'' in
  \emph{Proceedings of the 22nd ACM SIGSAC conference on computer and
  communications security}.\hskip 1em plus 0.5em minus 0.4em\relax ACM, 2015,
  pp. 1310--1321.

\bibitem{mcmahan2016communication}
H.~B. McMahan, E.~Moore, D.~Ramage, S.~Hampson \emph{et~al.},
  ``Communication-efficient learning of deep networks from decentralized
  data,'' \emph{arXiv preprint arXiv:1602.05629}, 2016.

\bibitem{wen2017terngrad}
W.~Wen, C.~Xu, F.~Yan, C.~Wu, Y.~Wang, Y.~Chen, and H.~Li, ``Terngrad: Ternary
  gradients to reduce communication in distributed deep learning,'' \emph{arXiv
  preprint arXiv:1705.07878}, 2017.

\bibitem{alistarh2017qsgd}
D.~Alistarh, D.~Grubic, J.~Li, R.~Tomioka, and M.~Vojnovic, ``Qsgd:
  Communication-efficient sgd via gradient quantization and encoding,'' in
  \emph{Advances in Neural Information Processing Systems}, 2017, pp.
  1707--1718.

\bibitem{seide20141}
F.~Seide, H.~Fu, J.~Droppo, G.~Li, and D.~Yu, ``1-bit stochastic gradient
  descent and its application to data-parallel distributed training of speech
  dnns,'' in \emph{Fifteenth Annual Conference of the International Speech
  Communication Association}, 2014.

\bibitem{bernstein2018signsgd}
J.~Bernstein, Y.-X. Wang, K.~Azizzadenesheli, and A.~Anandkumar, ``signsgd:
  compressed optimisation for non-convex problems,'' \emph{arXiv preprint
  arXiv:1802.04434}, 2018.

\bibitem{strom2015scalable}
N.~Strom, ``Scalable distributed dnn training using commodity gpu cloud
  computing,'' in \emph{Sixteenth Annual Conference of the International Speech
  Communication Association}, 2015.

\bibitem{aji2017sparse}
A.~F. Aji and K.~Heafield, ``Sparse communication for distributed gradient
  descent,'' \emph{arXiv preprint arXiv:1704.05021}, 2017.

\bibitem{lin2017deep}
Y.~Lin, S.~Han, H.~Mao, Y.~Wang, and W.~J. Dally, ``Deep gradient compression:
  Reducing the communication bandwidth for distributed training,'' \emph{arXiv
  preprint arXiv:1712.01887}, 2017.

\bibitem{konevcny2016federated}
J.~Kone{\v{c}}n{\`y}, H.~B. McMahan, F.~X. Yu, P.~Richt{\'a}rik, A.~T. Suresh,
  and D.~Bacon, ``Federated learning: Strategies for improving communication
  efficiency,'' \emph{arXiv preprint arXiv:1610.05492}, 2016.

\bibitem{golomb1966run}
S.~Golomb, ``Run-length encodings (corresp.),'' \emph{IEEE transactions on
  information theory}, vol.~12, no.~3, pp. 399--401, 1966.

\bibitem{lecun1998gradient}
Y.~LeCun, L.~Bottou, Y.~Bengio, and P.~Haffner, ``Gradient-based learning
  applied to document recognition,'' \emph{Proceedings of the IEEE}, vol.~86,
  no.~11, pp. 2278--2324, 1998.

\bibitem{lecun1998mnist}
Y.~LeCun, ``The mnist database of handwritten digits,'' \emph{http://yann.
  lecun. com/exdb/mnist/}, 1998.

\bibitem{krizhevsky2014cifar}
A.~Krizhevsky, V.~Nair, and G.~Hinton, ``The cifar-10 dataset,'' \emph{online:
  http://www. cs. toronto. edu/kriz/cifar. html}, 2014.

\bibitem{deng2009imagenet}
J.~Deng, W.~Dong, R.~Socher, L.-J. Li, K.~Li, and L.~Fei-Fei, ``Imagenet: A
  large-scale hierarchical image database,'' in \emph{Computer Vision and
  Pattern Recognition, 2009. CVPR 2009. IEEE Conference on}.\hskip 1em plus
  0.5em minus 0.4em\relax IEEE, 2009, pp. 248--255.

\bibitem{kingma2014adam}
D.~P. Kingma and J.~Ba, ``Adam: A method for stochastic optimization,''
  \emph{arXiv preprint arXiv:1412.6980}, 2014.

\bibitem{zaremba2014recurrent}
W.~Zaremba, I.~Sutskever, and O.~Vinyals, ``Recurrent neural network
  regularization,'' \emph{arXiv preprint arXiv:1409.2329}, 2014.

\bibitem{marcus1993building}
M.~P. Marcus, M.~A. Marcinkiewicz, and B.~Santorini, ``Building a large
  annotated corpus of english: The penn treebank,'' \emph{Computational
  linguistics}, vol.~19, no.~2, pp. 313--330, 1993.

\bibitem{shakespeare2014complete}
W.~Shakespeare, \emph{The complete works of William Shakespeare}.\hskip 1em
  plus 0.5em minus 0.4em\relax Race Point Publishing, 2014.

\bibitem{ioffe2015batch}
S.~Ioffe and C.~Szegedy, ``Batch normalization: Accelerating deep network
  training by reducing internal covariate shift,'' in \emph{International
  conference on machine learning}, 2015, pp. 448--456.

\bibitem{srivastava2014dropout}
N.~Srivastava, G.~Hinton, A.~Krizhevsky, I.~Sutskever, and R.~Salakhutdinov,
  ``Dropout: A simple way to prevent neural networks from overfitting,''
  \emph{The Journal of Machine Learning Research}, vol.~15, no.~1, pp.
  1929--1958, 2014.

\end{thebibliography}

\newpage
\section{Supplement}

\subsection{Momentum Correction, Warm-up Training and Momentum Masking:}
Lin et al. [17] introduce multiple minor modifications to the vanilla Gradient Dropping method. With these modifications they achieve up to around 1\% higher accuracy compared to Gradient Dropping on a variety of benchmarks. Those modifications include:

\emph{Momentum correction}: Instead of adding the raw gradient to the residuum, the momentum-corrected gradient is added. This is used implicitly in our approach, as our weight updates are already momentum-corrected.\\
\emph{Warm-up Training}: The sparsity rate is increased exponentially from 25\% to 0.1\% in the first epochs. We find that warm-up training can indeed speed-up convergence in the beginning of training, but ultimately has no effect on the final accuracy of the model. We therefore omit warm up training in our experiments, as it adds an additional hyperparameter to the method, without any real benefit.\\
\emph{Momentum Masking}: To avoid stale momentum from carrying the optimization into a wrong direction after a weight update is performed, Lin et al.\ suggest to set the momentum to zero for updated weights. We adopt momentum correction in our method.

\subsection{Golomb Position Decoding}
Algorithm \ref{alg:decode} describes the decoding of a binary sequence produced by Golomb Position Encoding (see main paper). Since the shapes of all weight-tensors are known to both the server and all clients, we can omit the shape information in both encoding and decoding.

\begin{algorithm}[H] \label{alg:decode}
\caption{Golomb Position Decoding}
\DontPrintSemicolon
\textbf{input:} binary message msg, bitsize $\mathbf{b}^*$, mean value $\mu$\\
\textbf{output:} sparse tensor $\Delta W^*$\\
\textbf{init:} $\Delta W^*\leftarrow 0\in \R^n$ \\
\textbullet~ $i\leftarrow0$; $q \leftarrow 0$; $j\leftarrow 0$\\
\While{$i<\text{size}(\msg)$}{
\uIf{$\msg[i]=0$}{
\textbullet~ $j\leftarrow j+q2^{\mathbf{b}^*}+\text{int}_{\mathbf{b}^*}(\msg[{i+1}],..,\msg[{i+\mathbf{b}^*}])+1$\\
\textbullet~ $\Delta W^*_j \leftarrow  \mu$\\
\textbullet~ $q\leftarrow 0$; $i\leftarrow i+\mathbf{b}^*+1$\\
}
\Else{
\textbullet~ $q\leftarrow q+1$; $i\leftarrow i+1$\\
}
}
\Return $\Delta W^*$
\end{algorithm}

\subsection{Model Specification}
Below, we describe the neural network models used in our experiments. Table \ref{tab:hyperparameters} list the training hyperparameters that were used.
\begin{table}[H]
\centering
\begin{tabular}{c|c|c|c|c|c}
& Iterations & Optimizer & Batchsize & LR & LR Decay\\
\hline
LeNet5-Caffe & 2000 & Adam [4] & 128$\times 4$ & 0.001 & -\\
ResNet32 & 60000 & Momentum @0.9 & 128$\times 4$ & 0.01 & 0.1 @ 30000, 50000\\
ResNet50 & 700000 & Momentum @0.9 & 32$\times 4$ & 0.1 & 0.1 @ 300000, 600000\\
WordLSTM & 60000 & Gradient Descent & 5$\times 4$ & 1.0 & 0.8 @ $24000+1200\times n$\\
CharLSTM & 16000 & Gradient Descent & 5$\times 4$ & 1.0 & 0.8 @ \makecell{5000, 8000, 10000,\\12000, 14000}
\end{tabular}
\caption{Hyperparameters used for our experiments in sections 3 and 4.}
\label{tab:hyperparameters}
\end{table}

\textbf{LeNet5-Caffe}:
The model specification can be downloaded from the Caffe MNIST tutorial page: \url{https://github.com/BVLC/caffe/blob/master/examples/mnist/lenet_train_test.prototxt}. (Features convolutional layers, fully connected layers, pooling.)\\

\textbf{ResNet32, ResNet50}: We use the implementation from the official Tensorflow repository: \url{https://github.com/tensorflow/models/tree/master/research/resnet}. (Features skip-connections, batch-normalization.)\\

\textbf{WordLSTM}: We use the implementation from the official Tensorflow repository (configuration "medium"): \url{https://github.com/tensorflow/models/tree/master/tutorials/rnn/ptb}. (Features trainable word-embeddings, multilayer LSTM-cells, dropout.)\\

\textbf{CharLSTM}: We adapt the implementation of WordLSTM to use a smaller vocabulary of 98 symbols by decreasing the size of the embedding. (Features trainable word-embeddings, multilayer LSTM-cells, dropout.)

\subsection{Proof of Theorem 2.1.}
\begin{proof}
It holds that
\begin{align}
\err(\mathcal{R}_{T-1}+\Delta W_T) = \|\sum_{t=1}^T\Delta W_t-\sum_{t=1}^{T-1}\Delta W_t^*-\mathcal{R}_{T-1}-\Delta W_T\|=0.
\end{align}
Since $\mathcal{S}$ is a metric subspace, the projection 
\begin{align}
\Delta W_T^* = \text{Proj}_\mathcal{S}(\mathcal{R}_{T-1}+\Delta W_T)
\end{align}
uniquely solves the minimization problem in $\mathcal{S}$.
\end{proof}

\subsection{Additional Results}

Figure \ref{fig:results_cifar} shows convergence speed in terms of iterations (left) and communicated bits (right) respectively for ResNet32 trained on CIFAR. Sparse Binary Compression can train the model to approximately baseline accuracy, using significantly less bits. SBC (3) trains the model to almost baseline accuracy (0.4\% degradation), while using $\times 32300$ less bits (cf. table 2). 

\begin{figure}[H]
\centering
\includegraphics[width=\textwidth]{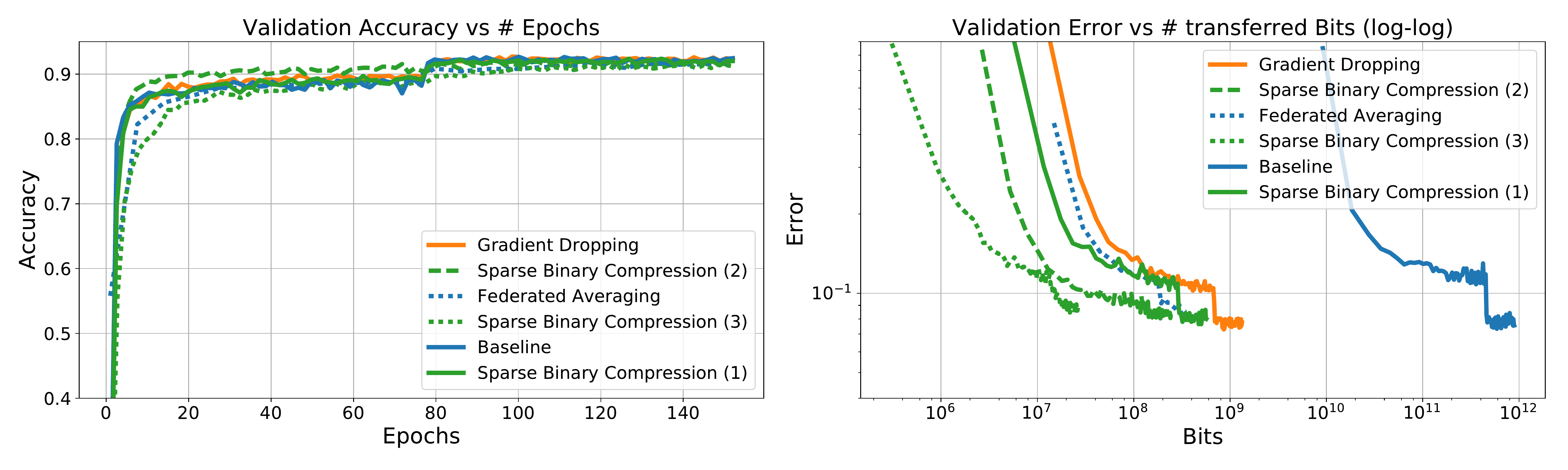}
\caption{Left: Top-1-Accuracy vs number of epochs. Right: Top-1-error vs transferred number of bits. Log-log plot. ResNet32 on CIFAR.}
\label{fig:results_cifar}
\end{figure}

Figure \ref{fig:results_charlstm} shows convergence speed in terms of iterations (left) and communicated bits (right) respectively for CharLSTM trained on Shakespeare. Sparse Binary Compression can train the model to approximately baseline accuracy, while using significantly less bits. SBC (1) even achieves the highest accuracy using $\times 2572$ less bits than the baseline (cf. table 2). SBC (3) however, shows non-monotonic convergence behavior on this benchmark. It might be that SBC (3) falls below the total communication budget necessary for this learning task (cf. section 3).

\begin{figure}[H]
\centering
\includegraphics[width=\textwidth]{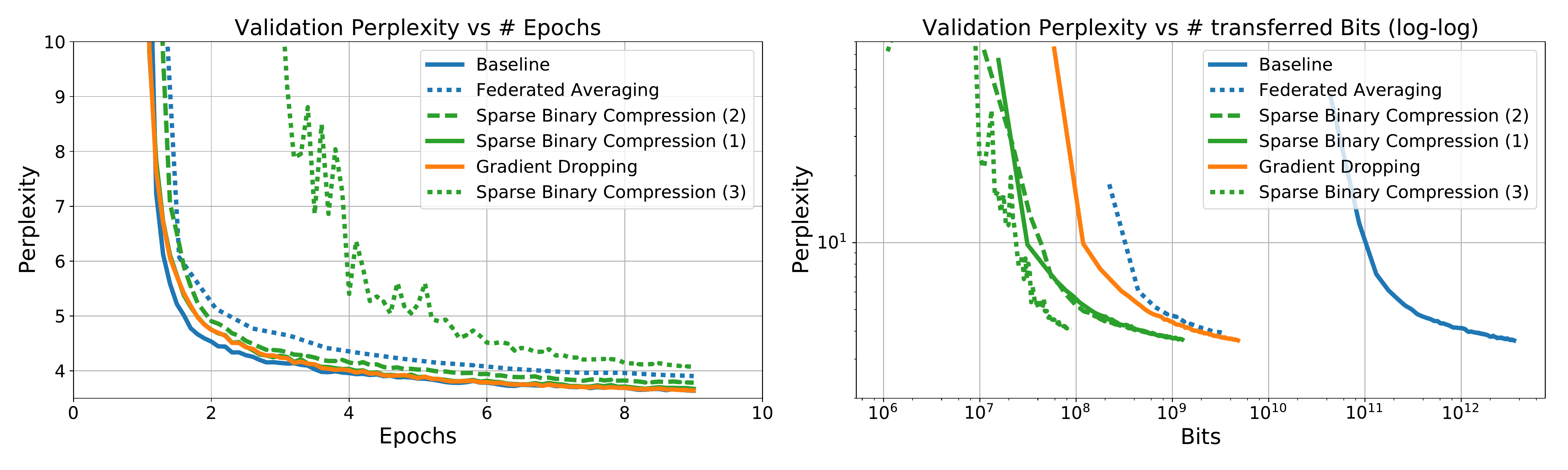}
\caption{Left: Perplexity vs number of epochs. Right: Perplexity vs transferred number of bits. Log-log plot. CharLSTM on Shakespeare.}
\label{fig:results_charlstm}
\end{figure}

Figure \ref{fig:ptb_appendix} shows validation error for WordLSTM trained on PTB at different levels of gradient sparsity and temporal sparsity. The total sparsity, defined as the product of temporal and gradient sparsity remains constant along the diagonals of the matrix. We observe that different forms of sparsity perform best during different stages of training. Phrased differently, this means that there is not one optimal sparsity setup, but rather sparsity needs to be adapted to the current training phase to achieve optimal compression.
\begin{figure}[H]
\centering
\includegraphics[width=\textwidth]{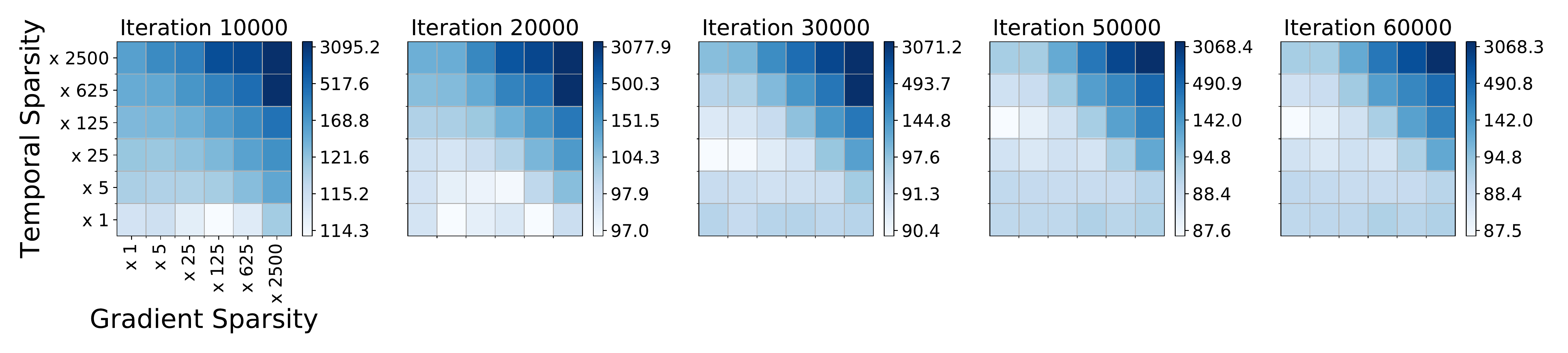}
\caption{Perplexity for different levels of gradient sparsity and temporal sparsity at different stages of training. WordLSTM trained on PTB.}
\label{fig:ptb_appendix}
\end{figure}

\end{document}